%% file: main.tex
\documentclass[letterpaper,10pt,conference]{ieeeconf}

\IEEEoverridecommandlockouts
\usepackage{mathptmx}
\usepackage{amsmath}
\usepackage{amssymb}
\usepackage{graphicx}
\usepackage{booktabs}
\usepackage{multirow}
\usepackage{pifont}
\usepackage{cite}
\usepackage[dvipsnames,table]{xcolor}
\usepackage[hidelinks]{hyperref}
\input{preamble}

\hypersetup{
  pdftitle={Stereo4DWalker: Learning 4D-aware Embodied Urban Navigation from Internet Stereo Videos},
  pdfauthor={Wentao Zhou, Xuweiyi Chen, Vignesh Rajagopal, Jeffrey Chen, Rohan Chandra, Zezhou Cheng}
}

\title{\LARGE \bf
Stereo4DWalker: Learning 4D-aware Embodied Urban Navigation \\
from Internet Stereo Videos
}

\author{Wentao Zhou, \space Xuweiyi Chen, \space Vignesh Rajagopal, \\
Jeffrey Chen, \space Rohan Chandra, \space Zezhou Cheng \\
University of Virginia \\
\url{https://www.cs.virginia.edu/~tsx4zn/stereowalk/}
}

\begin{document}

\maketitle
\thispagestyle{empty}
\pagestyle{empty}

\input{0_abstract}
\input{1_intro}

\input{2_related}
\input{2_5_dataset}
\input{3_method}
\input{4_experiments}
\input{5_conclusion}

\input{ack}

\bibliographystyle{IEEEtran}
\bibliography{main}

\end{document}

%% file: 0_abstract.tex
\begin{abstract}

Despite rapid progress, embodied navigation in dynamic and unstructured urban environments remains brittle. Most existing approaches directly map monocular visual inputs to actions through end-to-end pixel-to-action training, assuming that accurate spatiotemporal (4D) scene understanding will emerge implicitly. 
While appealing, this paradigm requires large amounts of pixel-to-action supervision that are difficult to obtain.
This challenge is amplified in dynamic, unstructured settings, where robust navigation requires precise 4D scene modeling.
To address these limitations, we present Stereo4DWalker, a 4D-aware embodied navigation model that leverages stereo inputs and  explicitly builds structured 4D representations of geometry and motion. 
These 4D structures are integrated into the navigation transformer through simple yet effective 4D-conditioned attention layers.
To support scalable training, we curate a large-scale stereo navigation dataset with automatically annotated actions from Internet stereo videos.
Our experiments show that Stereo4DWalker surpasses state-of-the-art performance using only 1.5\% of the training data, highlighting the effectiveness of explicit 4D visual modeling for data-efficient and robust urban navigation. 

\end{abstract}

%% file: 1_intro.tex
\section{Introduction}
\label{sec:intro}

Embodied urban navigation has attracted significant attention due to its potential for direct deployment in real-world applications such as last-mile delivery~\cite{liu2025citywalker,muhlbauer2009navigation,kummerle2013navigation,morales2009autonomous,mirowski2018learning}. 
Despite rapid progress, performance in dynamic and unstructured urban environments remains limited. 
Such environments involve dense pedestrian motion, irregular roadside layouts, and open-world object variability, posing substantial challenges to robust navigation. 
Effective operation in these settings requires accurate 4D scene understanding of geometry and dynamics, as well as adherence to common-sense and social conventions, including staying on sidewalks, obeying traffic signals, and maintaining appropriate interpersonal distance.

\input{figs/teaser}

Early visual navigation approaches~\cite{muhlbauer2009navigation,kummerle2013navigation,morales2009autonomous,guan2022ga,li2024stereonavnet} build on modular vision systems for detection, tracking, and planning, yet their performance was constrained by hard-coding rule-based decision making and evaluations confined to simple or near-static environments.
Reinforcement learning in photorealistic simulators~\cite{savva2019habitat,xia2018gibson, wu2025urbansim,xie2024vid2sim} achieved progress but still suffers from the sim-to-real gap.
Recent efforts~\cite{shah2023gnm,shah2023vint,sridhar2024nomad} introduced visual navigation foundation models that learn from expert demonstrations to map visual inputs directly to actions, however, learning these generalizable perception-action mapping remains fundamentally constrained by the lack of paired perception-action data at scale.

More recently, CityWalker~\cite{liu2025citywalker} addressed this limitation by automatically mining and annotating human navigation videos from the Internet. 
Although navigation models trained on in-the-wild Internet data have achieved impressive gains, their robustness in dynamic and unstructured environments remains limited for two reasons (Fig.~\ref{fig:teaser}a):
(1) they rely on monocular inputs, which introduce depth-scale ambiguity, amplify action-annotation noise, and impose fundamental performance bottlenecks;
$(2)$ 
existing urban navigation models directly map pixels to actions, assuming that geometric and dynamic understanding will emerge implicitly from supervision. This strategy requires substantial pixel-to-action annotations which are challenging to acquire in practice.
To address these limitations, we propose \textbf{Stereo4DWalker}, which integrates stereo vision and pretrained vision foundation models~\cite{karaev2025cotracker3,yang2024depth} into embodied navigation architectures to construct explicit 4D representations of scene geometry and motion.
Our work is guided by two key insights: 

First, \textbf{stereo inputs resolve the depth-scale ambiguity inherent in monocular perception.}
Although stereo cameras are widely adopted in robotics~\cite{kolter2009stereo,li2024stereonavnet}, their potential in urban navigation remains largely unexplored, primarily due to the absence of large-scale in-the-wild stereo navigation datasets with reliable action labels.
To address this issue, we develop an automated data collection pipeline that curates \textbf{StereoUrban}, a large-scale stereo navigation dataset from Internet videos. 
The pipeline leverages off-the-shelf vision-language models~\cite{wang2024qwen2} for data filtering and stereo odometry~\cite{qiu2025mac} for pseudo-label generation. 
The resulting dataset spans diverse scenes worldwide, complementing existing datasets that are limited in scale or restricted to monocular videos and simulation environments.
(Tab.~\ref{tab:dataset_comparison} and  Fig.~\ref{fig:data}).

\textbf{Second, integrating structured 4D scene understanding from vision foundation models into navigation architectures improves data efficiency and navigation performance.}
Recent advances in 4D scene understanding have been driven by highly generalizable and accurate vision foundation models~\cite{yang2024depth,lai2025tracktention,karaev2025cotracker3}. To incorporate their structured representations into navigation systems, we introduce simple yet effective 4D-conditioned attention layers (Fig.~\ref{fig:teaser}b). 
This module conditions the navigation transformer on explicit spatiotemporal structure, enabling control decisions to leverage metric geometry and dynamic consistency rather than relying solely on raw visual features. Experiments on existing urban navigation benchmarks and real-world deployment demonstrate that Stereo4DWalker surpasses state-of-the-art methods using only 1.5\% of the training data (Fig.~\ref{fig:teaser}c).

To summarize, our key contributions are: 
(1) We curate StereoUrban, a large-scale Internet stereo dataset of pedestrians across global metropolitan cities, together with an automated data filtering and scalable action annotation pipeline.
(2) We propose Stereo4DWalker, a 4D-aware embodied urban navigation framework that integrates explicit 4D scene structures from vision foundation models into a navigation transformer via a 4D-conditioned attention mechanism.
(3) We demonstrate that our model  design significantly improves data efficiency and navigation performance on established benchmarks and real-world deployments, underscoring the value of explicit 4D visual reasoning for robotic navigation.

%% file: figs/teaser.tex
\begin{figure}[!t]
    \centering
    \includegraphics[width=\linewidth]{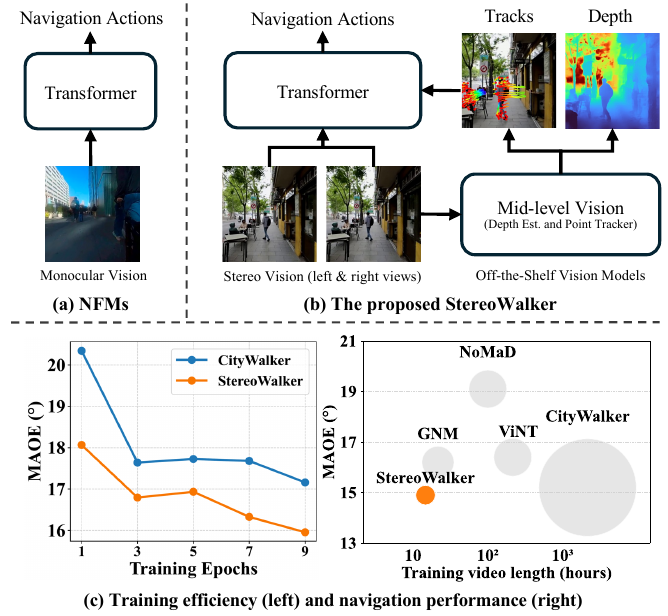}
    \vspace{-15pt}
    \caption{
        \textbf{Overview}. \textbf{(a)} Existing visual navigation foundation models (NFMs) process monocular inputs and predict actions without any intermediate vision modules. \textbf{(b)} StereoWalker improves this paradigm by incorporating stereo and mid-level vision modules such as depth estimation and dense point tracking. \textbf{(c)} Relative to CityWalker, the current state of the art, StereoWalker reaches high training efficiency and improved navigation accuracy while using only 1.5\% of the training data.
    }
    \label{fig:teaser}
\end{figure}

%% file: 2_related.tex
\section{Related Work}
\label{sec:related}

\noindent
\textbf{Embodied Visual Navigation.} 
Embodied navigation are commonly categorized by how the goal is specified, including point-goal~\cite{chaplot2020learning,savva2019habitat}, image-goal~\cite{zhu2017target,savinov2018semiparametric}, object-goal~\cite{chaplot2020object,majumdar2022zson}, and vision--language navigation~\cite{das2018embodied,krantz2020beyond}.
Our work follows the position-goal setting, where navigation is defined through ordered waypoints in Euclidean space.
A closely related line of work is \emph{social navigation}, which focuses on safe and socially acceptable motion around people by modeling human comfort and interaction dynamics~\cite{singamaneni2024survey}.
Previous navigation foundation models~\cite{shah2023gnm,shah2023vint,sridhar2024nomad} learn policies from large-scale demonstrations, while newer works extend this paradigm to in-the-wild video~\cite{hirose2024lelan,liu2025citywalker}, online self-improvement around pedestrians~\cite{hirose2024selfi}, and richer spatial reasoning with RGB-D input, memory, and web-scale human trajectories~\cite{liu2025vid,mei2025urbannav}.
Our setting overlaps with social navigation, as pedestrians and crowds are important sources of scene dynamics. However, rather than modeling social norms explicitly, we strengthen navigation foundation models through 4D representations, specifically stereo disparity, depth, and point tracking that many existing monocular-RGB approaches do not fully exploit.

\noindent
\textbf{Stereo and Mid-level Vision for Robotics.}
Prior work~\cite{chen2020robust,zhou2019does,sax2019learning,chen2025probing} has shown that intermediate visual representations (\eg, depth, semantic masks) improve generalization and data efficiency across robotic tasks.
Early studies~\cite{muller2018driving,mousavian2019visual} found that augmenting policies with cues such as optical flow, depth, and segmentation improves action prediction, and later work~\cite{zhou2019does} showed that such features often outperform raw pixels.
Recent navigation-oriented robotics works echo this trend: SELFI~\cite{hirose2024selfi} highlights the value of explicit adaptation around pedestrians, while D-CLING~\cite{nakaoka2025d} uses depth-conditioned fine-tuning to adapt navigation foundation models.
Stereo cameras provide a practical source of metric geometry with passive sensing, low cost, and simple deployment~\cite{kolter2009stereo,li2024stereonavnet}, and explicit 3D-temporal structure has also proven valuable for dynamic environments~\cite{zhang2024dynamic}.
However, these works do not study dynamic urban waypoint navigation from large-scale in-the-wild demonstrations.
In contrast, we show that explicitly incorporating stereo, depth, and dense tracking into urban navigation foundation models improves data efficiency and navigation accuracy.

\noindent
\textbf{Learning Robotics from Internet Videos.}
Large-scale Internet videos are increasingly being leveraged to reduce reliance on expensive robot-collected data for training robotic systems.
Prior work has explored transferable representation learning from passive videos~\cite{zakka2022xirl,nair2022r3m} or direct imitation from human videos despite embodiment mismatch~\cite{bahl2022human,kareer2025egomimic}.
While these methods show that videos provide scalable supervision for robotics, they mainly focus on manipulation or short-horizon control.
More recent works extend this idea to navigation.
For instance,
LeLaN~\cite{hirose2024lelan} learns language-conditioned navigation from in-the-wild videos. CityWalker~\cite{liu2025citywalker} scales urban navigation from web-scale monocular walking videos with pseudo-labels. 
Navigation World Model~\cite{bar2025navigation} studies controllable future prediction from egocentric video, 
and EgoWalk~\cite{akhtyamov2025egowalk} contributes a multimodal navigation dataset.
However, these approaches largely rely on monocular observations and do not explicitly exploit stereo geometry or structured 4D motion cues.
In contrast, our work studies \emph{stereo} urban navigation from in-the-wild VR180 videos and explicitly incorporates depth and dense tracking into the navigation model.

%% file: 2_5_dataset.tex
\section{Stereo Urban Navigation Dataset}
\label{sec:stereo-dataset}

\input{table/datasets}

While stereo sensing is widely used in robotic perception, it remains underexplored in embodied urban navigation, where most visual models rely on monocular or depth-simulated imagery. As depicted in Table.~\ref{tab:dataset_comparison}, existing datasets~\cite{nguyen2023musohu,scand,liu2025citywalker,bae2023sit,wu2025urbansim} often lack diverse stereo navigation data or rely on synthetic environments.
Fortunately, a large collection of stereoscopic city-walking videos is available on the Internet.
This unique data source enables us to curate a large-scale stereo navigation dataset.
Below we describe the key components of our data curation pipeline.

\textbf{Filtering and Quality Control.} 
Online walking footage often includes segments where the camera wearer pauses, interacts with bystanders, or engages in activities such as shopping or sightseeing. 
To remove these segments, we use a vision-language model, Qwen2-VL~\cite{wang2024qwen2}, to analyze visual content together with temporal context. 
For each candidate clip, the model reviews video frames, along with associated captions, and assesses whether the motion reflects forward locomotion toward an implicit goal. Only clips that consistently display egocentric, target-oriented walking are retained. 
The resulting dataset consists of continuous walking sequences with clear ego-motion and minimal idle behavior.

\textbf{Automatic Navigation Action Labeling.}
To derive action supervision for training our stereo navigation model, we compute trajectory labels directly from rectified stereo videos using state-of-the-art stereo visual odometry methods. 
In particular, we adopt MAC-VO~\cite{qiu2025mac}
on teleoperated navigation sequences with LiDAR-based SLAM ground truth. 
This automatic labeling pipeline processes large volumes of raw stereo footage without manual annotation or language-based prompting, enabling scalable expansion of training data.

We ultimately build \textbf{StereoUrban}, a stereo navigation dataset comprising 600 high-resolution stereo first-person walking clips totaling 60 hours of footage across multiple global cities, including San Francisco, Madrid, and Tokyo (Fig.~\ref{fig:data}). 
As described in Tab.~\ref{tab:dataset_comparison},
compared to existing navigation datasets, such as CityWalker~\cite{liu2025citywalker} which was collected within a single metropolitan region, StereoUrban offers substantially greater diversity in architectural layouts, lighting conditions, weather, and pedestrian density. We will release annotations, metadata, and video links under a CC license.

\input{figs/data}

%% file: table/datasets.tex
\begin{table}[t]
\vspace{5pt}
\centering
\caption{%
    \textbf{Comparison of embodied urban navigation datasets.}
    The proposed StereoUrban dataset provides the first large-scale, real-world stereo navigation benchmark for urban scenes collected from Internet videos.
}
\label{tab:dataset_comparison}
\setlength{\tabcolsep}{5pt}
\begin{tabular}{lccccc}
\toprule
Dataset & Diversity & Stereo & Real-world & \#Traj. & Source \\
\midrule
MuSoHu~\cite{nguyen2023musohu}        & \xmark & \xmark & \cmark & 300 & Manual \\
SCAND~\cite{scand}                    & \xmark & \xmark & \cmark & 138 & Manual \\
CityWalker~\cite{liu2025citywalker}   & \xmark & \xmark & \cmark & 598 & Internet \\
SiT~\cite{bae2023sit}                 & \xmark & \cmark & \cmark & 60  & Manual \\
UrbanSim~\cite{wu2025urbansim}        & \xmark & \cmark & \xmark & \textbackslash & Synthetic \\
\midrule
\rowcolor{gray!15}
\textbf{StereoUrban}               & \cmark & \cmark & \cmark & 600    & Internet \\
\bottomrule
\vspace{-28pt}
\end{tabular}
\end{table}

%% file: figs/data.tex
\begin{figure}[t]
    \vspace{5pt}
    \centering
    \includegraphics[width=\linewidth]{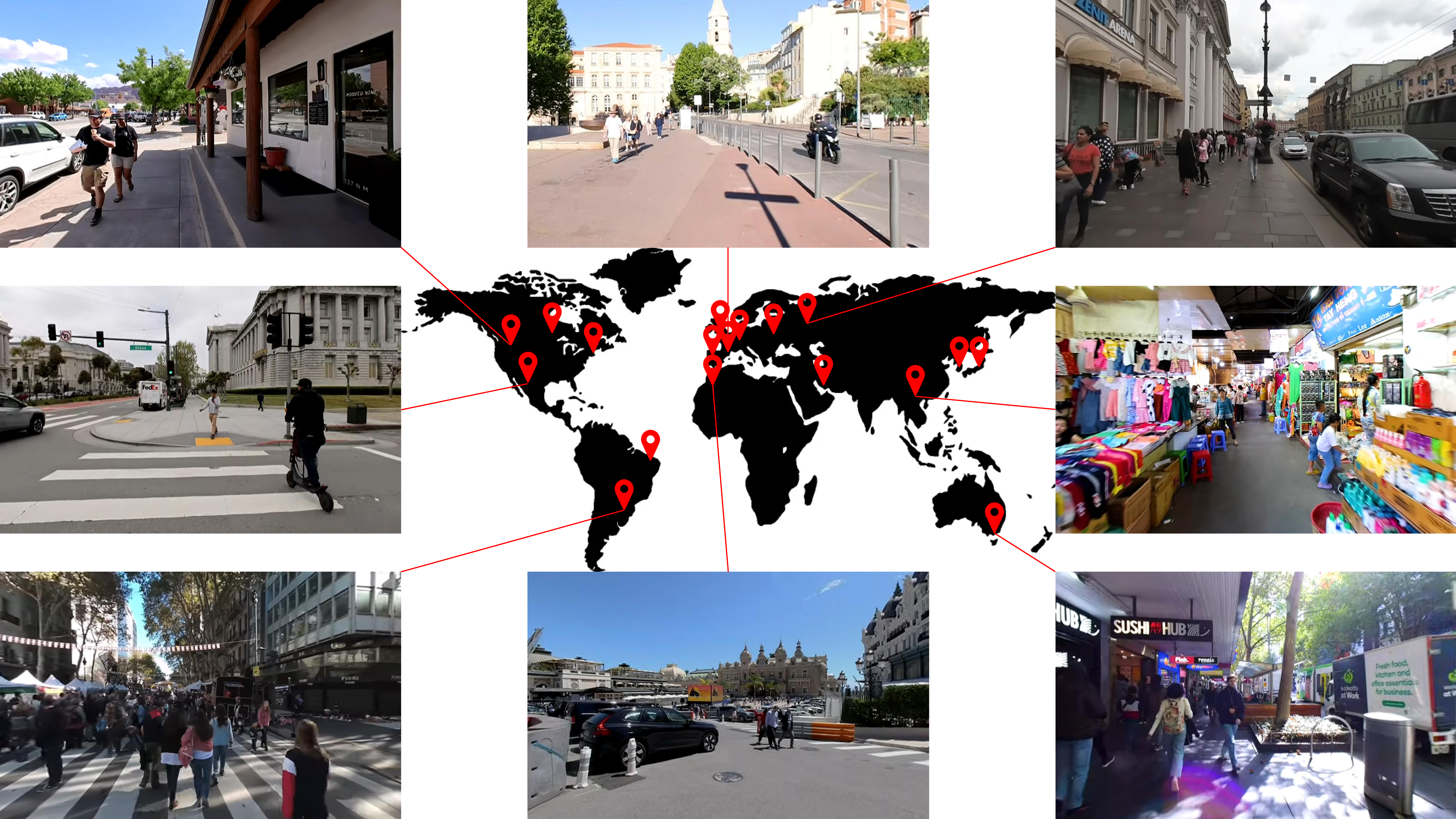}
    \caption{
        \textbf{Diversity of StereoUrban Dataset.} We curate a large-scale stereo navigation dataset from Internet city-walking videos, spanning diverse scenes worldwide.
    }
    \label{fig:data}
            \vspace{-20pt}

\end{figure}

%% file: 3_method.tex
\section{Stereo4DWalker}
\input{figs/main}

In this section, we first discuss the task setting of urban navigation (Sec.~\ref{subsec:pre}), and then provide the details of Stereo4DWalker (Sec.~\ref{subsec:Stereo4DWalker}).

\subsection{Preliminaries: Embodied Urban Navigation}
\label{subsec:pre}
We study embodied visual navigation in urban settings, where the objective is to generate a series of waypoints from the current location to a specified target.
Urban environments introduce substantial challenges due to their complex visual structure and human and vehicle motion, which demands an understanding of geometry and dynamics from raw visual input.
Following the task setup of previous urban navigation works~\cite{liu2025citywalker,sridhar2024nomad,shah2023vint,shah2023gnm}, 
each training instance consists of a temporally ordered sequence of visual observations $\left\{\mathcal{V}_i\right\}_{i=1}^N$, corresponding locations $\left\{\mathcal{P}_i \in \mathbb{R}^2\right\}_{i=1}^N$, and a sub-goal waypoint $\mathcal{P}' \in \mathbb{R}^2$. The learning objective is to approximate the following neural function:

{\small \[
    F_\theta: \left(\mathcal{V}_{t-N+1:t},\, \mathcal{P}_{t-N+1:t},\, \mathcal{P}'\right) \mapsto \mathcal{P}_{t+1:t+N},
\]}

where $N$ denotes the length of a short temporal window of recent stereo observations and positions. Our model takes the current subgoal $\mathcal{P}'$ as the immediate target and predicts the short-horizon trajectory that advances toward it. Once the model reaches $\mathcal{P}'$, the next waypoint in the predefined sequence becomes the new sub-goal, forming a continuous waypoint-to-waypoint navigation process. This formulation can naturally scale to long-range navigation by chaining multiple waypoints obtained from a global path planner such as $A^*$ and other graph-search based algorithms.

\subsection{Model Design}
\label{subsec:Stereo4DWalker}
Fig.~\ref{fig:pipeline} provides an overview of Stereo4DWalker architecture.
Given a short temporal window of rectified stereo (or monocular) frames
$\mathcal{V}_{t-N+1:t}$
and their corresponding positions 
$\mathcal{P}_{t-N+1:t}$,
the model forms dense image tokens that jointly encode appearance, geometry, and short-term motion cues. 
Tokens from all frames are then processed by three stages: $(i)$ 4D-conditioned attention layers to aggregate explicit 4D scene structure into navigation transformer, $(ii)$ global attention to integrate scene context across views, and $(iii)$ target-token attention to focus prediction on goal-relevant regions. 
Stereo4DWalker supports both stereo and monocular inputs with the same architecture.

\textbf{Image Tokenization.}
We use off-the-shelf vision foundation models to obtain 4D semantic, geometric and dynamic structures of the visual observations.
Specifically, we use DINOv2~\cite{oquab2023dinov2} for semantic-aware patch representations.
While prior visual navigation models~\cite{liu2025citywalker} compress each frame to a single DINOv2 \texttt{[CLS]} token, Stereo4DWalker retains \emph{all} patch tokens to preserve fine-grained spatial structure critical for control. 
Our intuition is straightforward: accurate navigation demands richer visual perception than a global summary can provide.

\textbf{4D-conditioned Attention Layers.}
For monocular input, 
we apply DepthAnythingV2~\cite{yang2024depth} to predict per-pixel depth, and MonSter++~\cite{cheng2025monster} for stereo alignment.
The depth map is then encoded via a simple CNN encoder into depth embeddings $\mathbf{z}_{j,k}$, with $j$ and $k$ indexing the spatial location of each image patch.
The geometric information is aggregated into the navigation transformer via a simple token concatenation strategy: concatenating the patch token $\mathbf{x}_{j,k}$ from DINOv2 and the depth embedding into $\mathbf{h}_{j,k} = \left[\,\mathbf{x}_{j,k};\,\mathbf{z}_{j,k}\,\right]$ (Fig.~\ref{fig:pipeline}b).

To capture temporal correspondence across frames, we introduce a \textit{track transformer} inspired by TrackTention~\cite{lai2025tracktention} (Fig.~\ref{fig:pipeline}b). 
This module integrates temporal motion into image features through three successive operations.

\textit{(1) Track integration:}
We first obtain dense pixel tracks from CoTracker3~\cite{karaev2025cotracker3}
within a temporal window $[t-N+1,t]$. 
Each point track is characterized by point coordinates into track tokens.
Using cross-attention between the track tokens and image tokens, the model aggregates local visual evidence from $\mathbf{h}_{j,k}$ into a set of track features $\mathbf{S}_{1:\mathbf{M}}$, where $\mathbf{M}$ denotes the number of tracks. 
This operation allows each track to selectively pool information from nearby spatial regions based on learned attention weights rather than fixed interpolation.

\textit{(2) Temporal attention:}
The track features $\mathbf{S}_{1:\mathbf{M}}$ from consecutive frames are organized into sequences of length $N$ for each track and processed by a transformer that performs self-attention along the temporal axis.
The resulting representations $\tilde{\mathbf{S}}_{1:\mathbf{M}}$ capture smooth temporal evolution of appearance, geometry, and motion.

\textit{(3) Feature update:}
To update image features, we perform a second cross-attention where patch coordinates, reflecting 2D spatial positions within the image, act as queries and the updated track tokens $\tilde{\mathbf{S}}_{1:\mathbf{M}}$ serve as keys and values.
This produces updated image tokens $\tilde{\mathbf{h}}_{j,k}$ that encode both 3D geometric structure and dynamic correspondence.

\textbf{Global Attention and Action Prediction.}
For navigation conditioning, the sub-goal waypoint $\mathcal{P}'$ and the recent trajectory $\mathcal{P}_{t-N+1:t}$ are projected through a shared lightweight MLP to produce a target token $\mathbf{g}$ and trajectory tokens, respectively.
After the track transformer, the image tokens $\tilde{\mathbf{h}}_{j,k}$, trajectory tokens, and target token $\mathbf{g}$ are concatenated into a unified sequence and jointly processed by multi-head self-attention layers, forming the \textit{global attention} stage. 
This yields an updated target token $\tilde{\mathbf{g}}$ that aggregates contextual information over the past $N$ seconds. 
Finally, $\tilde{\mathbf{g}}$ is fed into two MLP heads: an \textit{arrival head} that predicts the probability of reaching $\mathcal{P}'$, and an \textit{action head} that predicts the next $N$ waypoints $\mathcal{P}_{t+1:t+N}$.

\textbf{Training Objectives.}
During training, we minimize a composite loss which is formulated as
\begin{equation}
\mathcal{L}_{\text{total}} = 
\mathcal{L}_{\text{wp}} + 
\lambda_{\text{arrvd}} \mathcal{L}_{\text{arrvd}} + 
\lambda_{\text{dir}} \mathcal{L}_{\text{dir}}
\end{equation}
where $\mathcal{L}_{\text{wp}}$ denotes the waypoint prediction loss measuring spatial deviation between predicted and ground-truth trajectories, 
$\mathcal{L}_{\text{arrvd}}$ supervises the arrival probability at the target waypoint, 
$\mathcal{L}_{\text{dir}}$ compares the mean angle difference of each step. $\lambda_{\text{arrvd}}$ and $\lambda_{\text{dir}}$ are scalar weights balancing the auxiliary objectives, with $\lambda_{\text{arrvd}}=1.0$ and $\lambda_{\text{dir}}=10.0$.

%% file: figs/main.tex
\begin{figure*}[t]
    \centering
    \includegraphics[width=\linewidth]{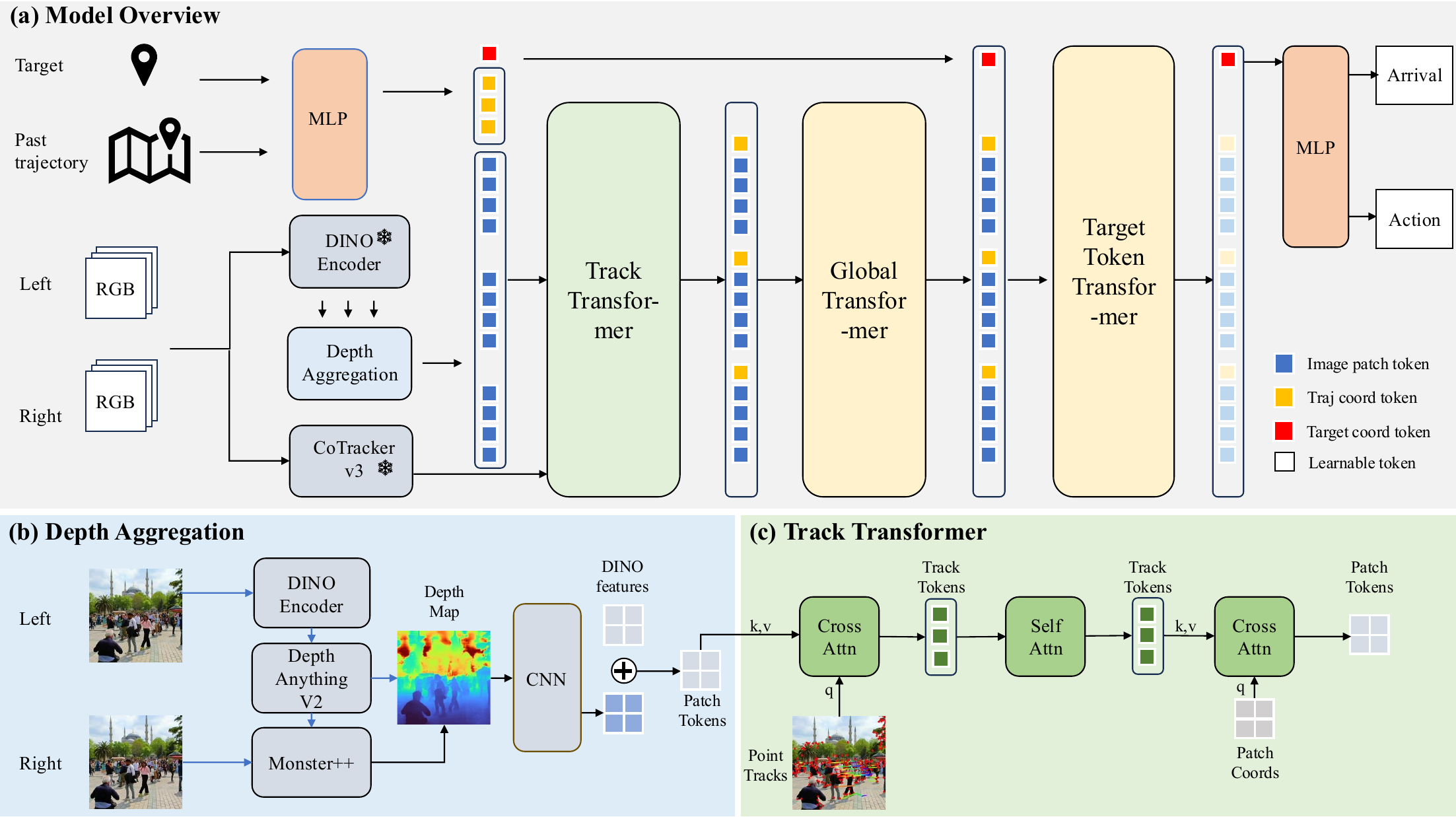}
    \caption{
        \textbf{Overview of StereoWalker's Architecture.}
        StereoWalker takes in stereo image pairs with corresponding trajectory and a target position. The images first go through three vision foundation modules, where depth embeddings and DINO features are concatenated to get patch tokens. We then append position tokens and use trackings for Tracktention~\cite{lai2025tracktention}. A global attention module further processes the tokens, followed by the target-token attention which merges all the information into the target token. Finally, an MLP converts the updated target token to arrival probability and action predictions. %
    }
    \label{fig:pipeline}
            \vspace{-10pt}

\end{figure*}

%% file: 4_experiments.tex
\section{Experiments}
\label{sec:exp}
In this section, we evaluate the performance of our proposed framework, Stereo4DWalker, for goal-directed navigation using monocular and stereo visual inputs. We aim to address the following key questions: $(i)$ does explicitly incorporating 4D visual representations improve robot navigation over state-of-the-art navigation frameworks? $(ii)$ does training with high quality stereo data enhance navigation robustness and accuracy compared to monocular setups? and $(iii)$ does Stereo4DWalker transfer reliably to real world robot deployments in a variety of critical navigation scenarios? To answer these questions, we outline the experimental setup, including baseline methods, evaluation metrics, and dataset collection, in Sec.~\ref{subsec:setup}. We then present stereo and monocular benchmarking results in Sec.~\ref{subsec:benchmark}. Finally, we provide detailed analysis of our results in Sec.~\ref{subsec:analysis}.

\input{table/citywalker_main}

\subsection{Experimental Setup}
\label{subsec:setup}

\textbf{Baselines.}
We compare our model with outdoor navigation models closely related to our setup, including GNM~\cite{shah2023gnm}, ViNT~\cite{shah2023vint}, NoMaD~\cite{sridhar2024nomad} and CityWalker~\cite{liu2025citywalker}. Although originally developed for image-goal navigation, we tested GNM~\cite{shah2023gnm}, ViNT~\cite{shah2023vint}, and NoMaD~\cite{sridhar2024nomad} using goal-images from our collected data. 
We also compare with indoor navigation model ViNL~\cite{kareer2023vinl} which takes in ground-truth depth. 
Without public pretrained weights, we train and test it with the same data as ours on monocular benchmark. 
We also considered StereoNavNet~\cite{li2024stereonavnet} which was originally trained in simulation environments, but its high computational cost made re-training on our urban benchmarks infeasible. 

\textbf{Evaluation Metrics.}
We evaluate predicted trajectories with two complementary metrics:
Maximum Average orientation error (MAOE), Arrival Accuracy and Euclidean distance. Following CityWalker~\cite{liu2025citywalker}, let $\theta_{i,k}$ denote the orientation (heading) error between the predicted motion direction and the ground-truth direction for sample $i$ at future step $k$.
With $N$ samples and a horizon of $T$ steps, we define

{\small \begin{equation}
\label{eq:maoe}
\mathrm{MAOE}
=\frac{1}{N}\sum_{i=1}^{N}\;\max_{1\le k\le T}\,\theta_{i,k}.
\end{equation}}

That is, for each sample we take the worst per-step orientation error over the horizon and then average across samples. MAOE captures the model's ability to maintain a correct heading toward the sub-goal, independent of trajectory scale. Let $\hat{\mathbf{p}}_{i,k}$ be the predicted position for sample $i$ at step $k$ and $\mathbf{g}_i$ the target waypoint.
Given a radius $r$ and a step budget $K$, the Arrival Accuracy is the fraction of samples that come within $r$ of the target within the next $K$ steps:

{\small \begin{equation}
\label{eq:arrival-acc}
\mathrm{Acc}_{\text{arr}}(r,K)
=\frac{1}{N}\sum_{i=1}^{N}\mathbf{1}\!\left[
\min_{1\le k\le K}\bigl\|\hat{\mathbf{p}}_{i,k}-\mathbf{g}_i\bigr\|_2 \le r
\right].
\end{equation}}

This metric directly reflects waypoint-level success and provides an interpretable notion of spatial goal attainment, with $r=5\,\mathrm{m}$ and $K=5$ in all experiments. Although Euclidean \(\ell_2\) distance is a straightforward measure of positional discrepancy, prior work~\cite{liu2025citywalker} shows it can miss navigation quality, \ie, trajectories may have small \(\ell_2\) error yet head in the wrong direction. Nevertheless, \(\ell_2\) remains a useful indicator of absolute localization accuracy and scale, so we report it alongside MAOE and Arrival Accuracy as a complementary metric.

\textbf{Expert Data.}
We collected expert demonstrations through teleoperation for fine-tuning and offline evaluation. The dataset was acquired using a Clearpath Jackal equipped with an Ouster LiDAR and a ZED 2i stereo camera. A LiDAR-based SLAM system was used to estimate the robot's pose, which served as the ground-truth action labels. Additionally, wheel odometry provided continuous relative motion estimates, enabling the robot to infer its current pose and predict its future trajectory. We also evaluated our model in a monocular setting using the CityWalker benchmark~\cite{liu2025citywalker}.

\subsection{Performance Benchmarking}
\label{subsec:benchmark}
We benchmark our model and strong baselines on the CityWalker benchmark and on our own stereo benchmark. In addition, we validate the findings with real-world navigation deployments.

\textbf{Monocular Benchmark.} \Cref{tab:benchmark_citywalker} (Monocular Benchmark) reports results across six critical scenarios: Turn, Crossing (Cross), Detour (Det.), Proximity (Prox.), Crowd (Crwd), and Other (Oth.). Overall, our fine-tuned model improves performance in most categories, improving MAOE by an average of 4--13\%, and arrival rates by 1--19\%. Fine-tuned Stereo4DWalker achieves the best scores on many metrics, in every scenario except the ``turn'' case. We hypothesize that the relative weakness on turns arises from data imbalance toward straight segments and the amplification of small orientation errors during sharp heading changes. Notably, explicitly adding mid-level vision features, including depth and tracking, improves performance on most categories, underscoring the value of structured geometric and temporal cues for monocular navigation. We provide more in depth analysis of each modality in Sec.~\ref{subsec:analysis}.

\input{table/real_robot}

\input{figs/visualization}

\textbf{Stereo Benchmark.} \Cref{tab:benchmark_citywalker} also summarizes results across critical scenarios on our offline Stereo4DWalker benchmark. Because our baselines do not natively ingest stereo, we feed their left image and also train a monocular variant of our model for a fair comparison. Even without stereo, our \emph{monocular} model already yields substantial gains, reducing average L2 error by 17--73\%, MAOE by 11--48\%, and arrival rates by 3--24\%, with the qualitative results shown in Fig.~\ref{fig:visualization}. 
Training with \emph{stereo} further achieves SOTA performance with consistent gains in \textit{Crossing}, \textit{Detour}, \textit{Crowd}, and \textit{Other}. The \textit{Turn} scenario remains challenging, with arrival lagging behind the monocular baselines despite comparable orientation error, suggesting the need for more turning-rich data or turn-aware objectives. Overall, stereo supervision provides a clear benefit beyond monocular baselines, reducing overall L2 error by 18--73\%, MAOE by 22--54\%, and arrival rates by 3--25\%.

\textbf{Real-world Deployment}
We evaluate our model in real-world setting using a Clearpath Jackal J100 robot.
The two strongest models on offline benchmarks, our model and CityWalker~\cite{liu2025citywalker}, are tested with identical start and end points to ensure comparability.
Each model runs remotely on a GPU server and communicates with the robot through FastAPI, which streams the predicted waypoints.
The low-level controller on ROS 2 converts the predicted short-horizon trajectory into continuous velocity commands for execution. 
We evaluate three motion patterns including forward, left turn, and right turn, with 14 trials conducted for each case.
A trial is considered successful when the robot arrives within 1m of the designated target and stays in that area; interruptions due to collisions are marked as failures. The consistent improvements across all motion types suggests that our model architecture provides more stable waypoint estimation under dynamic conditions, and the stereo training data boosts the performance further.

\subsection{Analysis}
\label{subsec:analysis}

\textbf{Analysis of 4D Vision.}
\Cref{tab:ablation} presents an ablation analysis that evaluates different architectural configurations on the CityWalker teleoperation benchmark. All variants are trained on the same monocular dataset, with specific components selectively enabled or disabled for a fair comparison. Earlier baselines such as ViNT~\cite{shah2023vint}, GNM~\cite{shah2023gnm}, NoMaD~\cite{sridhar2024nomad}, and CityWalker~\cite{liu2025citywalker} represent each image using only a single \texttt{[CLS]} token. In contrast, we observe that using all patch tokens to capture finer spatial information leads to an immediate 3.7\% improvement in the Mean Angular Orientation Error (MAOE). Building upon this representation, incorporating depth and dense pixel tracking further enhances navigation accuracy, as these two mid-level cues provide complementary inductive signals. Depth captures the three-dimensional structure, and adding depth yields a further 4.0\% reduction in MAOE relative to the patch token model.

Tracking encodes scene motion and temporal consistency, and incorporating tracking on top of patch tokens and depth provides an additional 2.8\% reduction in MAOE. Prior studies~\cite{chen2020robust,zhou2019does} demonstrated similar advantages of mid-level vision in controlled or static environments. Our experiments in large-scale dynamic urban navigation, showing that explicitly modeling depth and motion significantly improves robustness and effectiveness in real-world conditions, shown in Fig.~\ref{fig:real_world_overview}. Fine-tuned Stereo4DWalker improves performance by an average of 23.8\% over the Forward, Left turn, and Right turn scenarios. 
Each design takes 10 epochs and shares the same hyperparameter settings.

\begin{table}[t]
\vspace{5pt}
    \caption{\textbf{Ablation Study.} Effect of patch tokens, depth, and dense pixel tracking on navigation performance. We report the average MAOE ($^{\circ}$) -- lower is better.}
    \label{tab:ablation}
    \centering
    \resizebox{0.9\linewidth}{!}{
    \begin{tabular}{ccc|l}
    \toprule
        \textbf{Patch Tokens} & \textbf{Depth} & \textbf{Tracking} & \textbf{MAOE} ($\downarrow$) \\ 
    \midrule
        \xmark & \xmark & \xmark & 17.55 \\
        \cmark & \xmark & \xmark & 16.90 \textcolor{ForestGreen}{(-0.65)} \\
        \cmark & \xmark & \cmark & 16.40 \textcolor{ForestGreen}{(-1.15)} \\
        \cmark & \cmark & \xmark & 16.23 \textcolor{ForestGreen}{(-1.32)} \\
        \cmark & \cmark & \cmark & 15.77 \textcolor{ForestGreen}{(-1.78)}\\
    \bottomrule
    \end{tabular}
    }
    \vspace{-5mm}
\end{table}

\textbf{Analysis of Training Efficiency.}
\begin{figure}[t]
\vspace{5pt}
    \centering
    \includegraphics[width=\linewidth]{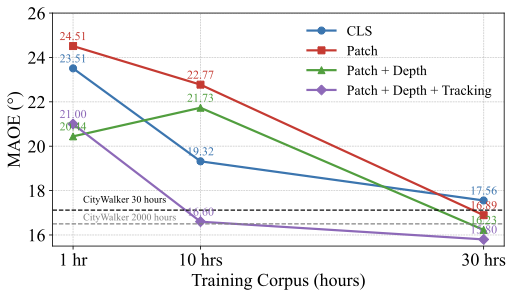}
    \vspace{-5mm}
    \caption{\textbf{Model Efficiency}. Our model empowered by \emph{mid-level vision capabilities} can surpass 2000 hours trained CityWalker using only 30 hours data with same epochs.}
    \vspace{-5mm}
    \label{fig:loss}
\end{figure}
Beyond performance gains, we also observe that incorporating explicit 4D visual structures significantly accelerates training, as depicted in Fig.~\ref{fig:loss}. We carefully train our model on monocular data using only a fraction of the original dataset, achieving comparable performance with merely 1.5\% of CityWalker's training data. Under different amounts of training data, we use the same training settings for both CityWalker and our model. Enabling patch tokens introduces richer visual representations but also necessitates architectural modifications to the decoder, as our model no longer relies on a single \texttt{[CLS]} token representation. Consequently, our model with patch tokens alone does not surpass CityWalker when trained for 30 hours. However, once depth cues are injected, the model already outperforms CityWalker. Further incorporating both depth and tracking information leads to faster convergence and superior performance, surpassing CityWalker trained with over 2,000 hours of monocular videos. This demonstrates that mid-level vision not only enhances representation quality but also provides strong inductive biases that make training more data- and time-efficient.

\textbf{Analysis of Inference Efficiency.}
Despite the incorporation of pretrained vision modules, the computational footprint of Stereo4DWalker remains within the practical range of modern robotic platforms. Specifically, our whole model requires 2.89\,GB of VRAM and 0.2\,s per sample on an A100 GPU at inference time, compared to CityWalker which consumes 1.68\,GB and 0.06\,s per sample.
This overhead remains manageable in our setting, as the model predicts five future waypoints spanning a five-second horizon. 
This allows the system to operate at a one-second inference interval, ensuring that new waypoint commands can be generated and executed between consecutive positions without causing stalls or instability in robot motion. 
As a result, even when accounting for network latency introduced by the FastAPI server, 
Stereo4DWalker remains capable of real-robots deployment and achieve the best performance.

%% file: table/citywalker_main.tex
\begin{table}[t]
    \centering
    \caption{\textbf{Benchmark on Offline Data}. We evaluate performance in monocular and stereo settings. The ``Mean'' column shows scenario means averaged over six scenarios; ``All'' shows sample means over all data samples. Best is \textbf{bold}; second-best is \underline{underlined}.\vspace{-2mm}}
    \label{tab:benchmark_citywalker}
    \setlength{\tabcolsep}{2.5pt} %
    \resizebox{\linewidth}{!}{
    \begin{tabular}{llcccccccc}
    \toprule
        \textbf{Method} & \textbf{Metric} & \textbf{Mean} & \textbf{Turn} & \textbf{Cross} & \textbf{Det.} & \textbf{Prox.} & \textbf{Crwd} & \textbf{Oth.} & \textbf{All} \\
        \midrule
        \multicolumn{10}{c}{\textit{Monocular Benchmark}} \\
        \midrule
        \multirow{3}{5em}{GNM~\cite{shah2023gnm}}
         & $\downarrow$ L2 (m) & 1.22 & 2.36 & 1.36 & 1.42 & \underline{0.88} & \underline{0.76} & \textbf{0.55} & \underline{0.74} \\
         & $\downarrow$ MAOE ($^{\circ}$) & 16.2 & 31.1 & 14.8 & \textbf{12.5} & 14.7 & 12.8 & 11.0 & 12.1 \\
         & $\uparrow$ Arrival (\%) & 68.6 & 66.4 & 69.7 & 66.4 & 69.0 & 69.2 & 70.7 & 70.0 \\
        \midrule
        \multirow{3}{5em}{ViNT~\cite{shah2023vint}}
         & $\downarrow$ L2 (m) & 1.30 & 1.91 & 1.13 & \underline{1.14} & \textbf{0.77} & \textbf{0.66} & \underline{0.57} & \textbf{0.70} \\
         & $\downarrow$ MAOE ($^{\circ}$) & 16.5 & 31.1 & 15.4 & \underline{12.9} & 14.8 & 13.3 & 11.6 & 12.6 \\
         & $\uparrow$ Arrival (\%) & 70.5 & \textbf{71.2} & 68.7 & 70.7 & 73.0 & 68.6 & 71.0 & 70.7 \\
        \midrule
        \multirow{3}{6em}{NoMaD~\cite{sridhar2024nomad}}
         & $\downarrow$ L2 (m) & 1.39 & 2.49 & 1.56 & 1.55 & 1.06 & 0.95 & 0.76 & \underline{0.74} \\
         & $\downarrow$ MAOE ($^{\circ}$) & 19.1 & 35.1 & 18.5 & 15.6 & 18.1 & 14.3 & 12.8 & 12.1 \\
         & $\uparrow$ Arrival (\%) & 68.6 & 66.4 & 69.7 & 66.4 & 69.0 & 69.3 & 70.7 & 70.0 \\
        \midrule
        \multirow{3}{7.5em}{Citywalker~\cite{liu2025citywalker}}
         & $\downarrow$ L2 (m) & 1.11 & \textbf{1.27} & \textbf{1.00} & 1.15 & 1.06 & 1.12 & 1.06 & 1.07 \\
         & $\downarrow$ MAOE ($^{\circ}$) & \underline{15.2} & \underline{26.6} & \underline{14.1} & 13.9 & \underline{14.3} & \underline{12.0} & 10.4 & \underline{11.5} \\
         & $\uparrow$ Arrival (\%) & \underline{81.8} & 68.9 & \underline{75.3} & 78.5 & \textbf{90.6} & 87.5 & 90.2 & 87.8 \\
        \midrule
        \multirow{3}{6em}{\textbf{StereoWalker} (zero-shot)}
         & $\downarrow$ L2 (m) & \underline{1.10} & \underline{1.30} & 1.03 & 1.24 & 1.01 & 1.04 & 0.98 & 1.02 \\
         & $\downarrow$ MAOE ($^{\circ}$) & 15.8 & \textbf{26.1} & 14.8 & 17.0 & 14.4 & 12.2 & \underline{10.2} & 11.7 \\
         & $\uparrow$ Arrival (\%) & 81.0 & 65.3 & 74.9 & \underline{78.9} & 86.5 & \underline{89.4} & \underline{90.9} & \underline{88.1} \\
        \midrule
        \multirow{3}{6em}{\textbf{StereoWalker}} 
         & $\downarrow$ L2 (m) & \textbf{1.06} & 1.37 & \underline{1.01} & \textbf{1.04} & 1.00 & 0.98 & 0.98 & 1.00 \\
         & $\downarrow$ MAOE ($^{\circ}$) & \textbf{14.9} & 28.1 & \textbf{13.5} & 13.0 & \textbf{14.0} & \textbf{11.2} & \textbf{9.8} & \textbf{11.0} \\
         & $\uparrow$ Arrival (\%) & \textbf{82.8} & \underline{70.0} & \textbf{77.9} & \textbf{80.2} & \underline{87.3} & \textbf{89.8} & \textbf{91.4} & \textbf{88.9} \\
        \midrule
        \multicolumn{10}{c}{\textit{Stereo Benchmark}} \\
        \midrule
        \multirow{3}{5em}{GNM~\cite{shah2023gnm}}
         & $\downarrow$ L2 (m) & 1.43 & 2.15 & 0.99 & 2.42 & 0.89 & 1.04 & 1.10 & 1.13 \\
         & $\downarrow$ MAOE ($^{\circ}$) & 8.0 & 24.8 & 4.4 & \underline{5.6} & 4.7 & 3.7 & 4.8 & 4.7 \\
         & $\uparrow$ Arrival (\%) & 68.8 & \textbf{75.0} & 70.5 & 61.3 & 70.0 & 65.2 & 70.6 & 69.4 \\
        \midrule
        \multirow{3}{5em}{ViNT~\cite{shah2023vint}}
         & $\downarrow$ L2 (m) & 1.79 & 2.53 & 1.31 & 3.11 & 0.99 & 1.38 & 1.45 & 1.47 \\
         & $\downarrow$ MAOE ($^{\circ}$) & 7.2 & 24.5 & 3.8 & 4.6 & 3.4 & 2.8 & 3.9 & 3.8 \\
         & $\uparrow$ Arrival (\%) & 71.4 & \textbf{75.0} & 68.8 & 80.6 & 60.0 & 73.8 & 70.3 & 70.6 \\
        \midrule
        \multirow{3}{6em}{NoMaD~\cite{sridhar2024nomad}}
         & $\downarrow$ L2 (m) & 2.28 & 2.12 & 2.23 & 2.74 & 2.02 & 2.24 & 2.31 & 2.31 \\
         & $\downarrow$ MAOE ($^{\circ}$) & 9.9 & 26.7 & 6.6 & 9.4 & 5.2 & 5.0 & 6.3 & 6.3 \\
         & $\uparrow$ Arrival (\%) & 72.2 & 66.7 & 68.4 & 71.0 & 80.0 & 75.2 & 72.0 & 71.7 \\
        \midrule
        \multirow{3}{7.5em}{CityWalker~\cite{liu2025citywalker}}
         & $\downarrow$ L2 (m) & 0.75 & \textbf{1.03} & 0.76 & \textbf{0.39} & 0.74 & 0.77 & 0.78 & 0.76 \\
         & $\downarrow$ MAOE ($^{\circ}$) & 5.8 & 15.4 & \underline{3.4} & 6.2 & 3.3 & 3.2 & 3.7 & 3.7 \\
         & $\uparrow$ Arrival (\%) & \underline{83.1} & \textbf{75.0} & 92.9 & 54.8 & \textbf{90.0} & 92.9 & 92.9 & 91.2 \\
        \midrule
        \multirow{3}{6em}{\textbf{StereoWalker} (monocular)}
         & $\downarrow$ L2 (m) & \textbf{0.69} & \underline{1.23} & \underline{0.66} & 0.55 & \textbf{0.48} & \underline{0.61} & \underline{0.63} & \underline{0.63} \\
         & $\downarrow$ MAOE ($^{\circ}$) & \underline{5.4} & \underline{14.6} & 3.6 & 5.9 & \textbf{2.6} & \underline{2.3} & \underline{3.3} & \underline{3.3} \\
         & $\uparrow$ Arrival (\%) & 82.0 & 62.5 & \textbf{95.5} & 51.6 & \textbf{90.0} & \textbf{96.5} & \underline{95.7} & \underline{93.7} \\
        \midrule
        \multirow{3}{6em}{\textbf{StereoWalker} (stereo)}
         & $\downarrow$ L2 (m) & \textbf{0.69} & 1.36 & \textbf{0.65} & \underline{0.42} & \textbf{0.48} & \textbf{0.59} & \textbf{0.63} & \textbf{0.62} \\
         & $\downarrow$ MAOE ($^{\circ}$) & \textbf{5.1} & \textbf{14.5} & \textbf{3.1} & \textbf{5.5} & \underline{2.7} & \textbf{2.0} & \textbf{3.0} & \textbf{2.9} \\
         & $\uparrow$ Arrival (\%) & \textbf{86.4} & 62.5 & \underline{93.8} & \textbf{80.6} & \textbf{90.0} & 95.0 & \textbf{96.2} & \textbf{94.6} \\
    \bottomrule
    \end{tabular}
    }
\end{table}

%% file: table/real_robot.tex
\begin{figure}[t]
  \centering
  \begin{minipage}[t]{0.47\columnwidth}
  \vspace{0pt}
    \centering
    \setlength{\tabcolsep}{4pt}
    \small
    \resizebox{\linewidth}{!}{%
      \renewcommand{\arraystretch}{1.5} 
        \begin{tabular}{lccccc}
        \toprule
        \textbf{Method} & \textbf{All} & \textbf{F} & \textbf{LT} & \textbf{RT} \\
        \midrule
        CW$^{\ast}$~\cite{liu2025citywalker} &50.0  &57.1  &42.9 &50.0 \\
        \midrule
        \textbf{Ours$^{\ast}$}              &66.7  &\textbf{85.7 } &64.3  &50.0  \\
        \midrule
        \textbf{Ours$^{\dagger}$}           &\textbf{73.8}  &\textbf{85.7}  &\textbf{71.4}  &\textbf{64.3}  \\
        \bottomrule
        \end{tabular}
    }
  \end{minipage}
  \hfill
  \begin{minipage}[t]{0.47\columnwidth}
  \vspace{0pt}
    \centering
    \includegraphics[width=\linewidth]{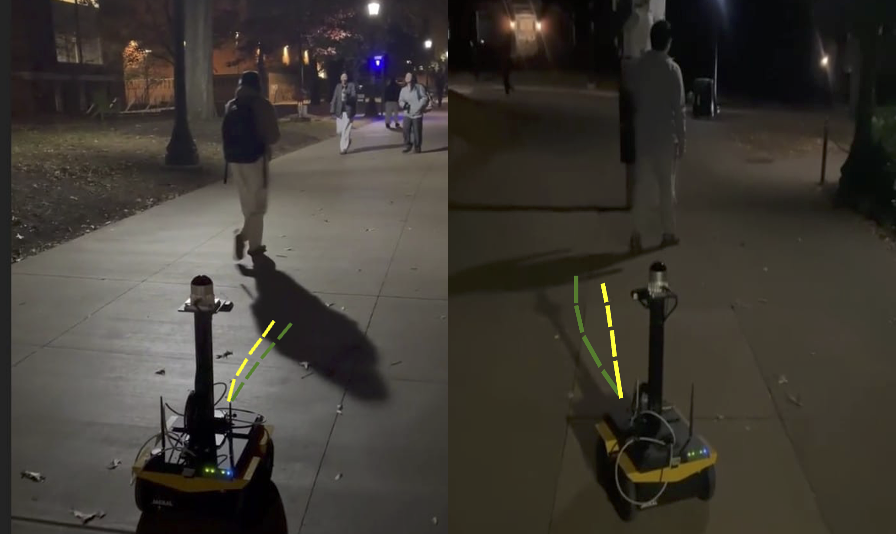}
  \end{minipage}
  \vspace{-3mm}
  \setlength{\abovecaptionskip}{3pt}
  \caption{\textbf{Real-World Navigation.} The table shows success rates (\%) for real-world experiments in different scenarios. $^\ast$ and $^\dagger$ indicate zero-shot inference and fine-tuned models, respectively. CW denotes CityWalker; F, LT, and RT denote Forward, Left Turn, and Right Turn. The images on the right compare StereoWalker (green) with CityWalker (yellow). StereoWalker maintains a safer distance from nearby pedestrians than CityWalker.}
  \label{fig:real_world_overview}
  \vspace{-15pt}
\end{figure}

%% file: figs/visualization.tex
\begin{figure*}[!t]
    \centering
    \includegraphics[width=\linewidth]{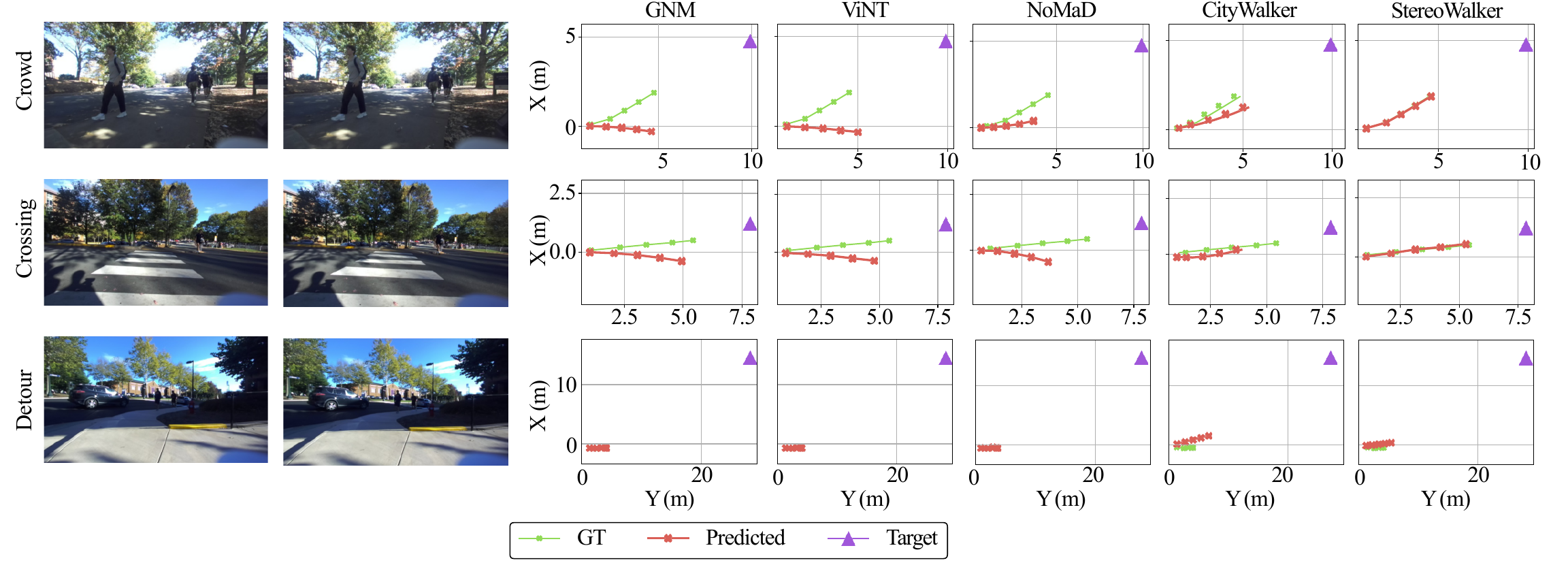}
    \vspace{-5mm}
    \caption{
        \textbf{Qualitative results.} We sample three trajectories from different critical scenarios in StereoWalker benchmark. In each case, the leftmost image depicts the current observation, followed by the predicted future waypoints visualized from different models. %
    }
    \label{fig:visualization}
    \vspace{-10pt}

\end{figure*}

%% file: 5_conclusion.tex
\section{Conclusion, Limitations, and Future Work}
\label{sec:conclusion}

\vspace{-1mm}
\noindent\paragraph*{Summary}
In this work, we present Stereo4DWalker, an embodied navigation transformer that integrates stereo inputs with explicit 4D vision modules for dynamic urban environments. 
To support this model, we collect a large-scale stereo videos from Internet city-walking footage, paired with an automatic filtering and action annotation process. 
Across curated benchmarks and real-world tests, our model achieves state-of-the-art performance with remarkably training efficiency. 
We further provide a detailed analysis on the effectiveness of stereo and 4D vision, highlighting the continued relevance of core computer vision representations in the development of urban robotic navigation models.

\noindent\paragraph*{Limitations} 
While Stereo4DWalker demonstrates the utility of stereo and 4D vision for embodied navigation, these ideas are not yet fully explored across the broader robotic tasks, ranging from mobile manipulators to aerial robots. 
Moreover, we aim to make the data curation pipeline more scalable, enabling larger and more diverse datasets with lower processing overhead.

%% file: ack.tex
\section*{Acknowledgment}

The authors acknowledge the Adobe Research Gift, the University of Virginia Research Computing and Data Analytics Center, Advanced Micro Devices AI and HPC Cluster Program, Advanced Cyberinfrastructure Coordination Ecosystem: Services \& Support (ACCESS) program, and National Artificial Intelligence Research Resource (NAIRR) Pilot for computational resources, including the Anvil supercomputer (National Science Foundation award OAC 2005632) at Purdue University and the Delta and DeltaAI advanced computing resources (National Science Foundation award OAC 2005572). PI Chandra is supported in part by a CCI award supported by the Commonwealth of Virginia.